%% file: manuscript.tex
\documentclass[letterpaper]{article} 
\usepackage[preprint]{aaai2027}  
\usepackage[hyphens]{url}  
\usepackage{graphicx} 
\usepackage{natbib}  
\usepackage{caption} 
\usepackage{algorithm}
\usepackage{algorithmic}

\usepackage{newfloat}
\usepackage{listings}
\DeclareCaptionStyle{ruled}{labelfont=normalfont,labelsep=colon,strut=off} 
\floatstyle{ruled}
\newfloat{listing}{tb}{lst}{}
\floatname{listing}{Listing}

\usepackage{booktabs}

\usepackage{amsmath}
\usepackage{amsfonts}
\usepackage[dvipsnames]{xcolor}
\usepackage{csquotes}
\usepackage{tcolorbox}
\tcbuselibrary{breakable}
\tcbuselibrary{skins}
\usepackage{enumitem}
\newtcolorbox{promptbox}[1][]{
    title={#1},
    colframe=gray!15,
    colbacktitle=gray!15,
    colback=gray!5,
    breakable=true,
    fonttitle=\bfseries\color{black},
}
\usepackage{booktabs}
\usepackage{multirow}
\usepackage{arydshln}
\usepackage{tabularray}
\usepackage{adjustbox}
\usepackage{xspace}
\usepackage{subcaption}
\usepackage{enumitem}
\usepackage{pgfplots}
\pgfplotsset{compat=1.18}
\usepackage{hyperref}

\newcommand{\best}[1]{\textbf{#1}}
\newcommand{\second}[1]{\underline{#1}}

\newcommand{\method}{RealWeather\xspace}
\title{\method: Realistic and Scene-Faithful Weather Translation with Driving World Models}
\author{
    Yuwei Ning\textsuperscript{\rm 1} \quad
    Liangzhi Wang\textsuperscript{\rm 1} \quad
    Yi Xiao\textsuperscript{\rm 1} \quad
    Zhenhua Wu\textsuperscript{\rm 1} \quad \\
    Yun Pang\textsuperscript{\rm 1} \quad
    Mingkun Chang\textsuperscript{\rm 1} \quad
    Jichang Li\textsuperscript{\rm 1} \quad
    Guanbin Li\textsuperscript{\rm 1, \rm 2}\corresponding
}
\affiliations{
    \textsuperscript{\rm 1}Sun Yat-sen University  \textsuperscript{\rm 2}Shenzhen Loop Area Institute \\
    {\tt\small
        ningyw@mail2.sysu.edu.cn, wanglzh26@mail2.sysu.edu.cn, xiaoy2622935705@gmail.com, \\
        wuzhh56@mail2.sysu.edu.cn, pangy9@mail2.sysu.edu.cn, mingkun502@gmail.com, \\
        lijichang@foxmail.com, liguanbin@mail.sysu.edu.cn
    }
}

\makeatletter
\newcommand{\aaaisteaser}{%
  \par
  \begingroup
    \centering
    \vspace*{-0.5em}
    \includegraphics[width=\textwidth]{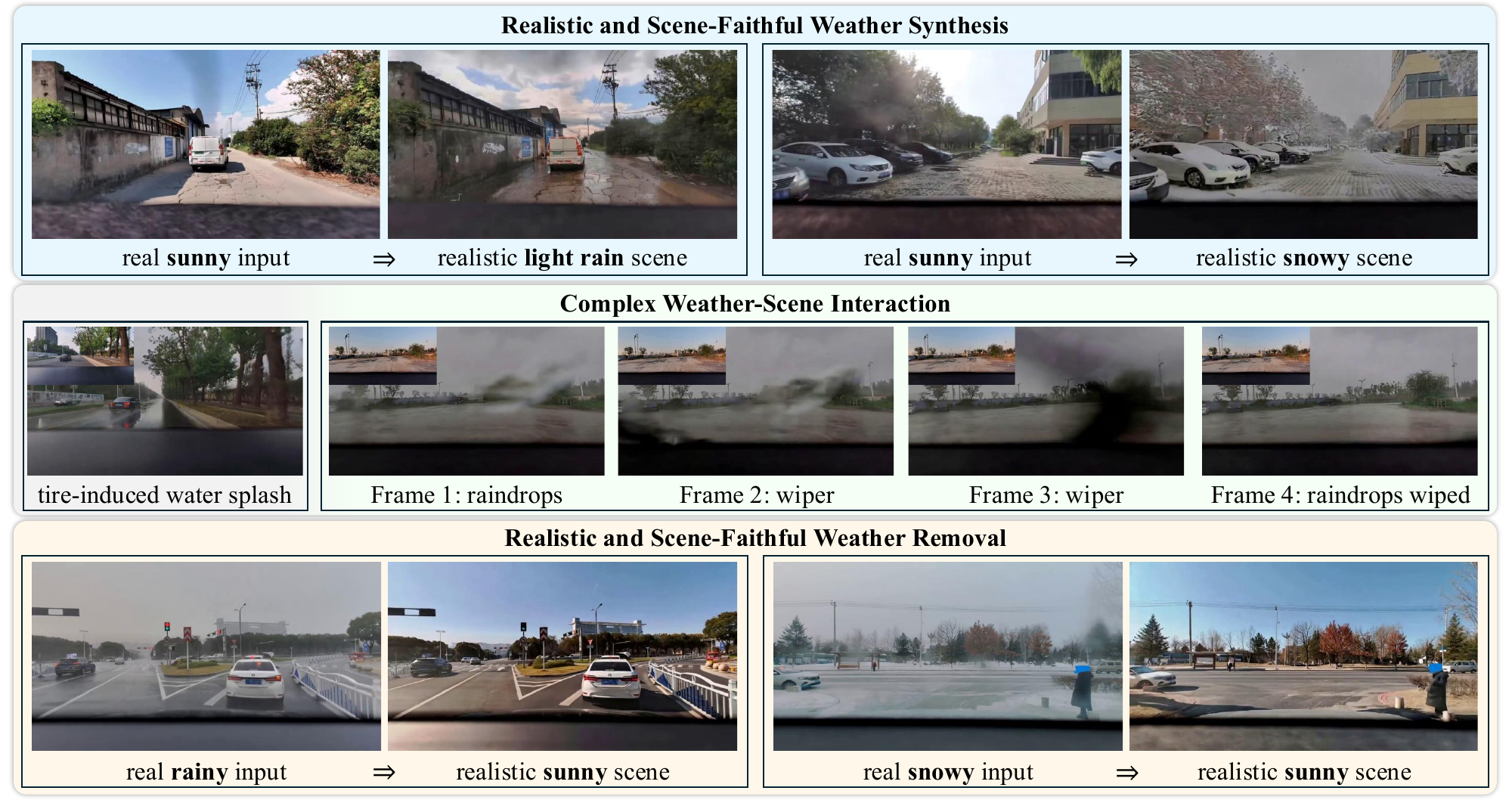}\par
    \vspace*{-0.5em}
    \captionof{figure}{%
        \textbf{\method} enables bidirectional, realistic, and scene-faithful weather translation for driving videos. Given a sunny driving video, \method synthesizes realistic adverse weather with complex dynamic effects, such as tire splashes and windshield-wiper interactions, while faithfully preserving road textures, vehicle appearance, and traffic layouts. It also supports the reverse translation, removing adverse-weather effects to recover clear scenes while preserving the underlying environment.
    }%
    \label{fig:teaser}%
  \endgroup
  \par
  \medskip
}
\g@addto@macro\@maketitle{\aaaisteaser}
\makeatother

\begin{document}
\maketitle 

\begin{abstract}
Realistic weather translation is valuable for developing and evaluating autonomous driving systems, yet collecting paired videos of the same scenes under different weather conditions at scale is impractical. Existing methods therefore rely on synthetic data, 3D weather editing, or geometry-conditioned generation, often compromising weather realism or scene fidelity. 
We propose \textbf{\method}, a driving world model for both realistic and scene-faithful weather translation. 
Our key idea is to learn authentic weather dynamics directly from real-world videos. Specifically, \method employs \textit{Progressive Realism Bootstrapping}, an iterative data-refinement strategy. Assisted by an auxiliary \textit{Pseudo-Clear Generation} pipeline, training initially starts with pseudo-style conditioning videos. As training proceeds, these inputs are progressively replaced with increasingly realistic videos generated by the model itself. This strategy bridges the pseudo-to-real domain gap, allowing the model to adapt seamlessly to real-world input distributions and naturally support bidirectional clear~$\leftrightarrow$~adverse translation. 
Furthermore, to strictly enforce structural integrity and suppress hallucinations, we introduce \textit{Scene-Fidelity RL Optimization}, a reward-driven policy optimization strategy that explicitly penalizes alterations to safety-critical driving elements. 
Extensive experiments demonstrate that \method significantly outperforms existing methods in visual realism and structural preservation, while enabling robust long-tail weather scenario generation and strong zero-shot out-of-distribution generalization.
Our video demos can be found at \href{https://hust-umi.github.io/RealWeather/}{this URL}.
\end{abstract}

\section{Introduction}

Ensuring the safety of autonomous driving systems requires rigorous training and evaluation across diverse, long-tail adverse weather conditions~\cite{hu2023_uniad, zeng2026futuresightdrive, dauner2024navsim, weng2024drive}. Since collecting and annotating real-world driving data under extreme weather is prohibitively expensive and unscalable, generative driving world models have emerged as a promising alternative for data synthesis. Specifically, video-to-video weather translation is highly desirable: it promises to expose perception and planning models to rare weather scenarios while inherently preserving the original traffic dynamics and scene layouts~\cite{liao2025diffusiondrive, xu2026wod}. Crucially, this translation must be strictly scene-faithful; any structural hallucinations (e.g., altering lanes or vehicle identities) would invalidate the original driving labels, rendering the synthesized data useless for downstream tasks.

However, achieving weather translation that is both highly realistic (capturing complex physical interactions like wet-road reflections) and strictly scene-faithful remains a formidable challenge. Existing approaches mainly follow three paradigms: 1) \textit{Synthetic-data-driven methods}~\cite{yin2026holoworld, zhu2026intrinsicweather, lin2025controllable, kanlis2024synthrsf} rely on paired data from graphics engines or simulators; 2) \textit{3D-based methods}~\cite{qian2026weathervid, qian2026weatheredit, wu2026weathercity, liu2026autoweather4d} render precipitation through handcrafted particle simulation and background editing; and 3) \textit{Geometry-conditioned generative models}~\cite{hu2026autoawg, ali2025world, chen2025unimlvg} use structural cues (e.g., depth maps) with text prompts to guide translation. Unfortunately, these methods suffer from synthetic-to-real domain gaps, rigid handcrafted rules, or a lack of appearance cues to preserve fine-grained details and synthesize weather-induced structures. In short, existing methods force an unacceptable compromise between realistic weather rendering and faithful scene preservation.

In this work, we present \textbf{\method}, a driving world model that achieves both realistic and scene-faithful weather translation. Instead of relying on synthetic supervision or handcrafted simulation, our key insight is that \textit{real-world adverse-weather videos themselves provide the most reliable supervision for learning authentic weather dynamics}, provided an effective training strategy can overcome the lack of strictly paired data.
To realize this, we propose \textbf{Progressive Realism Bootstrapping}, an iterative data-refinement strategy. Training starts with real adverse-weather targets paired with initially pseudo-clear conditioning videos. Because these pseudo-style inputs inevitably fail to capture real-world imaging effects (e.g., motion blur, sunlight glare), we progressively replace them with increasingly realistic videos generated by the model itself. As training proceeds, these refined inputs bridge the pseudo-to-real domain gap, allowing the model to adapt seamlessly to real-world input distributions and support bidirectional clear~$\leftrightarrow$~adverse translation.

Beyond realistic weather effects, explicitly enforcing structural integrity is imperative to preserve original driving labels. We introduce \textbf{Scene-Fidelity RL Optimization}, a reward-driven policy optimization strategy. By incorporating segmentation-based rewards on safety-critical elements (e.g., lane markings, vehicles, and pedestrians), it strictly penalizes structural hallucinations. This ensures that \method maintains precise scene layouts and object identities without compromising its powerfully learned weather editing capability.
Additionally, to overcome the unpaired data dilemma and kickstart this training loop, we develop an auxiliary \textbf{Pseudo-Clear Generation} pipeline that converts real adverse-weather videos into temporally coherent pseudo-clear counterparts, establishing the foundational translation capability required for bootstrapping.

Extensive evaluations on the Waymo Open Dataset, nuScenes, and a custom driving dataset demonstrate that \method generates significantly more realistic and scene-faithful translations than existing approaches. As shown in Figure~\ref{fig:teaser}, our model faithfully captures complex physical interactions, including windshield wiping, water splashes, wet-road reflections, and snow accumulation. Furthermore, \method enables robust long-tail scenario generation and shows strong zero-shot generalization to day-to-night translation and fisheye camera views. 
The main contributions of \method are summarized as follows:
\begin{itemize}[leftmargin=*, topsep=0pt, itemsep=2pt]
    \item We propose \textbf{\method}, a driving world model that learns realistic and scene-faithful bidirectional weather translation directly from real-world videos. Assisted by an auxiliary \textbf{pseudo-clear generation} pipeline, \method eliminates the dependence on synthetic paired supervision.
    \item We introduce a dual-optimization strategy comprising \textbf{Progressive Realism Bootstrapping} to bridge the pseudo-to-real input gap and enable bidirectional translation, alongside \textbf{Scene-Fidelity RL Optimization} to explicitly suppress structural hallucinations.
    \item Extensive experiments on benchmarks show that \method outperforms state-of-the-art methods in visual realism and structural preservation, demonstrating strong capabilities in long-tail weather synthesis and zero-shot out-of-distribution generalization.
\end{itemize}

\begin{figure*}[t!]
    \centering
    \includegraphics[width=1.0\linewidth]{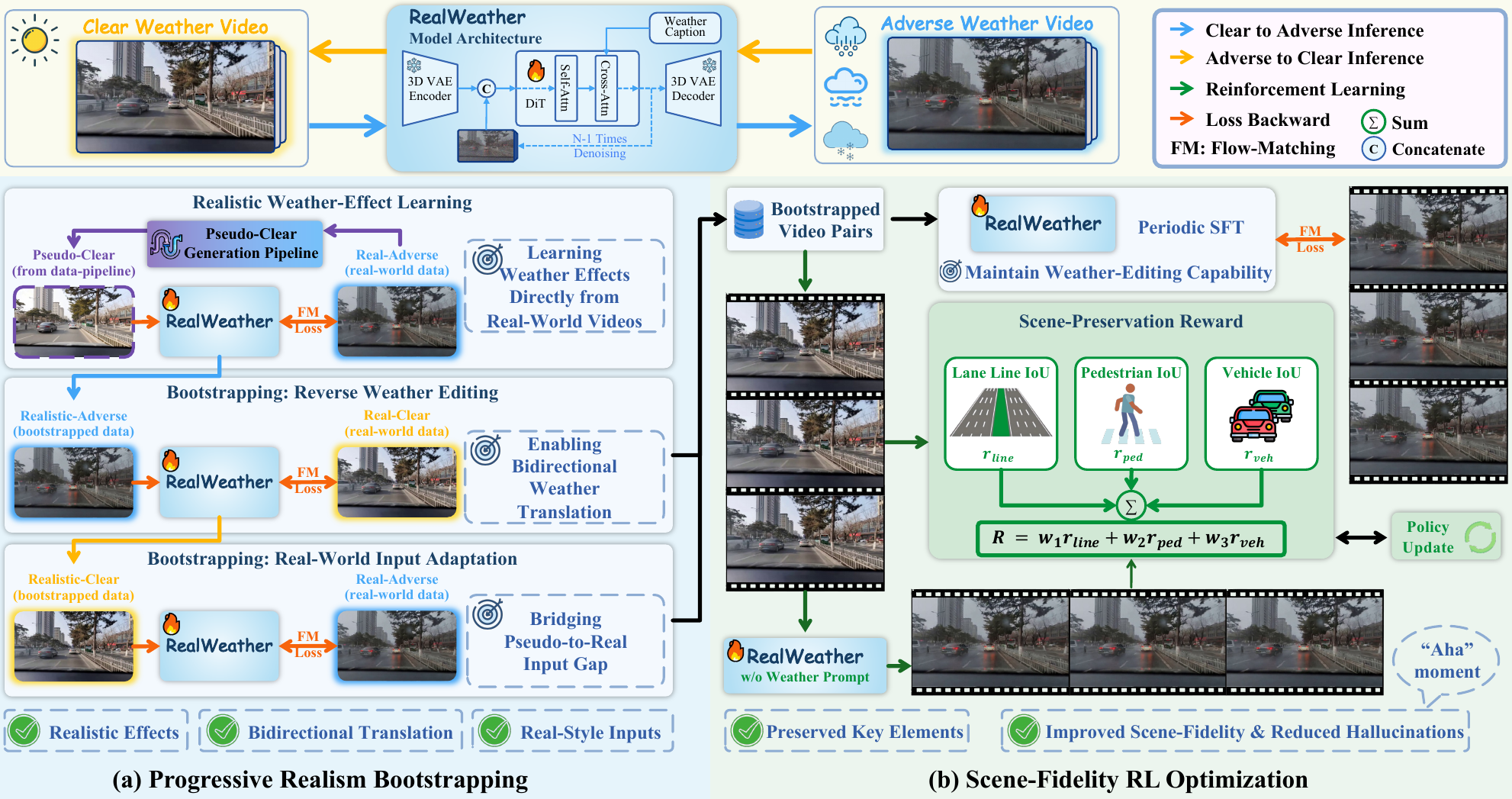}
    \caption{
        Overview of \textbf{\method}.
        (A) \textit{Progressive Realism Bootstrapping} enables the model to learn weather effects directly from real-world videos, progressively adapt to real-style inputs, and support bidirectional weather editing.
        (B) \textit{Scene-Fidelity RL Optimization} further improves structural fidelity with segmentation-based rewards, reducing hallucinations while preserving the learned weather translation capability.
    }

    \label{fig:method}
\end{figure*}

\section{Related Work}
\paragraph{Weather Removal and Synthesis.}
Early adverse-weather restoration methods mainly focus on removing a specific type of weather degradation, such as rain streaks, raindrops, snowflakes, or haze~\cite{wu2023ridcp, wu2024rainmamba, lai2025snowmaster, gao2024efficient}. These task-specific methods are later extended to all-in-one weather removal frameworks that aim to handle multiple degradations within a unified model~\cite{lu2025continuous, rajagopalan2025awracle}. Although these methods are effective for low-level restoration, their goal is primarily to remove transient weather artifacts and recover clean images. They usually do not model scene-level weather changes, such as snow accumulation or wet-road reflections.

Recent works move beyond weather removal and study controllable weather synthesis or translation. One line of work~\cite{yin2026holoworld, zhu2026intrinsicweather, lin2025controllable, kanlis2024synthrsf} collects paired clear/adverse training data using graphics engines~\cite{unrealengine, unity, godot} or driving simulators~\cite{Dosovitskiy17, li2022metadrive, airsim2017fsr}, and trains weather translation models on the synthetic pairs. 
Another line~\cite{qian2026weathervid, qian2026weatheredit, wu2026weathercity, liu2026autoweather4d} typically place handcrafted rain-and-snow particles in 3D space and render weather-affected videos in conjunction with weather-aware background editing.
These methods provide controllable weather type and intensity, but their reliance on synthetic assets, physical engines, or handcrafted weather rules often limits visual realism.
Although some methods~\cite{zhu2026intrinsicweather, lin2025controllable} incorporate real weather data by generating counterparts with models trained on simulator-generated pairs, they still largely rely on synthetic supervision to learn weather effects and struggle to adapt to real-weather input styles.
In contrast, our method learns weather effects directly from real driving videos and progressively improves the realism of training inputs during realism bootstrapping.

\paragraph{Generative Driving World Models.}
Recent generative driving world models can be broadly categorized into future-rollout models and geometry-conditioned generative models.
Future-rollout models synthesize future driving scenes from historical observations and future conditions, such as ego actions, camera trajectories, or text prompts~\cite{russell2025gaia, gao2024vista, wang2025mila, zhang2025epona}. These models can produce realistic driving scenarios with coherent temporal evolution.
Geometry-conditioned models use structural cues such as edges, depth maps, BEV layouts, or HD maps to guide driving scene generation~\cite{gao2025magicdrive, zhao2025drivedreamer, li2025uniscene, ali2025world}. With additional weather or style prompts, they can synthesize driving videos under different weather conditions. However, geometry-only conditions have two limitations for weather translation. First, clear-weather geometry may not match adverse-weather geometry, since snow accumulation and puddle reflections can introduce new geometry structures. Second, geometric cues do not capture detailed appearance, making it difficult to preserve vehicle colors, surface textures, and scene details.
In contrast, our method adapts a future-rollout driving world model into a video-to-video weather editor. With realism bootstrapping and scene-fidelity RL optimization, the adapted model achieves realistic and scene-faithful weather translation.

\section{Methodology}

\subsection{Overview of \method}

In this section, we introduce \textbf{\method}, a driving world model adapted for realistic and scene-faithful weather translation in driving videos. As illustrated in Figure~\ref{fig:method}, our framework orchestrates a dual-optimization strategy supported by an auxiliary data pipeline. Specifically, we first present \textbf{Progressive Realism Bootstrapping}, our core training mechanism that learns authentic weather dynamics from real-world videos and iteratively bridges the pseudo-to-real input domain gap. 
To explicitly enforce structural integrity, we then introduce \textbf{Scene-Fidelity RL Optimization}, a reward-driven policy optimization strategy that strictly penalizes structural hallucinations. 
Finally, to overcome the initial unpaired data dilemma and provide the necessary counterparts to kickstart the aforementioned training loop, we detail the auxiliary \textbf{Pseudo-Clear Generation} pipeline.

\paragraph{Model Architecture.}
\method is built upon a pretrained world model~(Cosmos-Predict2.5~\cite{ali2025world}) that generates driving videos with a flow-matching objective~\cite{lipman2022flow, liu2022flow}. Given a video $\mathbf{x}_0$, a 3D VAE~\cite{kingma2013auto} encodes it into a latent representation $\mathbf{z}_0=\mathcal{E}(\mathbf{x}_0)$. For flow matching, given Gaussian noise $\boldsymbol{\epsilon}\sim\mathcal{N}(0,\mathbf{I})$ and timestep $t\in[0,1]$, the interpolated latent is defined as
$
\mathbf{z}_t=t\boldsymbol{\epsilon}+(1-t)\mathbf{z}_0.
$
The pretrained model predicts the velocity field
$
{d\mathbf{z}_t}/{dt} \approx f_\theta(\mathbf{z}_t,t,c),
$
conditioned on the latent state $\mathbf{z}_t$, timestep $t$, and text prompt $c$.

To adapt the world model for weather translation, we additionally condition it on an input driving video $\mathbf{x}^{\mathrm{in}}$. Specifically, we encode the input video as $\mathbf{z}^{\mathrm{in}}=\mathcal{E}(\mathbf{x}^{\mathrm{in}})$ and concatenate it with the noisy target latent along the channel dimension to predict the velocity:
$
{d\mathbf{z}_t}/{dt} \approx f_\theta\left([\mathbf{z}_t,\mathbf{z}^{\mathrm{in}}],t,c\right).
$
During inference, we start from Gaussian noise and integrate the learned velocity field from $t=1$ to $t=0$:
\begin{equation} \label{eq:ode-sampling}
\mathbf{z}_{\mathrm{gen}}
=
\mathbf{z}_1
+
\int_{1}^{0}
f_\theta\left([\mathbf{z}_t,\mathbf{z}^{\mathrm{in}}],t,c\right)
\,dt,\ \mathbf{z}_1\sim\mathcal{N}(0,\mathbf{I}).
\end{equation}
The translated video is decoded by the VAE decoder $\mathcal{D}$ as
\begin{equation}
\mathbf{x}_{\mathrm{gen}}
=
\mathcal{D}\left(\mathbf{z}_{\mathrm{gen}}\right).
\end{equation}

\subsection{Progressive Realism Bootstrapping} \label{sec:training}

\paragraph{Realistic Weather-Effect Learning.}
Unlike previous methods that rely on synthetic supervision or handcrafted weather simulation, \method learns realistic weather effects from real-world adverse-weather videos. To enable paired supervision, we construct a pseudo-clear counterpart for each real-world adverse-weather video using the auxiliary \textit{Pseudo-Clear Generation} pipeline $\mathcal{P}_{\mathrm{pc}}$ (detailed subsequently).
Given a real-world adverse-weather video $x^{\mathrm{ra}}$ from our collected dataset $\mathcal{A}$, we generate its pseudo-clear counterpart $x^{\mathrm{pc}}$ as
\begin{equation}
x^\textrm{pc} = \mathcal{P}_{\mathrm{pc}}(x^\textrm{ra}).
\end{equation}
We then encode the pseudo-clear and real-world adverse-weather videos as
\begin{equation}
\mathbf{z}^\textrm{pc} = \mathcal{E}( x^\textrm{pc} ), 
\quad
\mathbf{z}^\textrm{ra} = \mathcal{E}( x^\textrm{ra} ).
\end{equation}
Following the flow-matching formulation, given Gaussian noise $\boldsymbol\epsilon \sim \mathcal{N}(0,\mathbf{I})$ and timestep $t \in [0,1]$, we construct the noisy target latent
\begin{equation}
\mathbf{z}^\textrm{ra}_t = t\boldsymbol\epsilon + (1-t)\mathbf{z}^\textrm{ra}.
\end{equation}
Then, the model learns realistic weather effects from real-world driving video dataset $\mathcal{A}$ under the following objective:
\begin{equation}
\mathcal{L}_{\mathrm{ra}} = \mathbb{E}_{
    x^\textrm{ra} \sim \mathcal{A}, \,
    \boldsymbol\epsilon, \,
    t
}[\|
f_\theta([\mathbf{z}^\textrm{ra}_t, \mathbf{z}^{\mathrm{pc}}],t,c_{a})
-
(\boldsymbol\epsilon-\mathbf{z}^{\mathrm{ra}})
\|_2^2],
\end{equation}
where $c_a$ denotes the target adverse-weather prompt corresponding to $x^{\mathrm{ra}}$.
This real-world supervised training provides the foundation for realistic weather-effect learning. 
However, the pseudo-clear inputs $x^{\mathrm{pc}}$ still differ from real-world clear-weather videos in imaging style, such as sunlight glare and camera-motion blur, leading to a \textit{pseudo-to-real input gap}. Moreover, the supervision above only covers the clear-to-adverse direction.
These limitations motivate the progressive bootstrapping mechanism introduced next, which adapts the model to real-style inputs and extends it to clear~\(\leftrightarrow\)~adverse bidirectional weather editing.

\paragraph{Bootstrapping: Reverse Weather Editing.}
Once initialized with realistic weather-effect learning, the model can synthesize realistic adverse-weather videos from pseudo-clear inputs. We leverage this capability to bootstrap training for the reverse editing direction. Specifically, given a real-world clear-weather video $x^{\mathrm{rc}}$ from the clear-weather dataset $\mathcal{C}$, we first process it using the same generation pipeline to match the pseudo-clear style:
\begin{equation}
x^\mathrm{rc \rightarrow pc} = \mathcal{P}_\mathrm{pc}(x^\mathrm{rc}).
\end{equation}
We then use the initialized model $f_{\theta_\mathrm{ra}}$ to translate the pseudo-clear-style video into a realistic adverse-weather video:
\begin{equation}
x^\mathrm{ba} = \mathcal{S}_{\theta_\mathrm{ra}}(x^\mathrm{rc \rightarrow pc}, c_a),
\end{equation}
where $x^{\mathrm{ba}}$ denotes the bootstrapped adverse-weather video, $c_a$ is the adverse-weather prompt, and $\mathcal{S}_{\theta_\mathrm{ra}}$ denotes the sampling process in Eq.~\ref{eq:ode-sampling} with the initialized model. The generated video $x^\mathrm{ba}$ is paired with the original real clear-weather video $x^\mathrm{rc}$, forming a bootstrapped reverse-editing pair $(x^\mathrm{ba}, x^\mathrm{rc})$.
We train the reverse editing direction with the flow-matching objective:
\begin{equation}
\mathcal{L}_\mathrm{re}
=
\mathbb{E}_{x^\mathrm{rc} \sim \mathcal{C}, \, \boldsymbol\epsilon, \, t}
[\|
f_\theta\left([\mathbf{z}^{\mathrm{rc}}_t,\mathbf{z}^{\mathrm{ba}}],t,c_c\right)
-
(\boldsymbol{\epsilon}-\mathbf{z}^{\mathrm{rc}})
\|_2^2],
\end{equation}
where $\mathbf{z}^{\mathrm{ba}}=\mathcal{E}(x^{\mathrm{ba}})$, $\mathbf{z}^{\mathrm{rc}}=\mathcal{E}(x^{\mathrm{rc}})$, $\mathbf{z}^{\mathrm{rc}}_t=t\boldsymbol{\epsilon}+(1-t)\mathbf{z}^{\mathrm{rc}}$, and $c_c$ denotes the clear-weather prompt.

\paragraph{Bootstrapping: Real-World Input Adaptation.}
Although reverse weather editing equips the model with adverse-to-clear translation ability, the forward direction is still initialized with pseudo-clear inputs, whose imaging style may differ from real clear-weather videos. To better adapt the model to real-world inputs, we use the reverse-editing model $f_{\theta_\mathrm{re}}$ to generate more realistic clear-weather counterparts for real adverse-weather videos. Given a real adverse-weather video $x^\mathrm{ra} \in \mathcal{A}$, we obtain a bootstrapped clear-weather video by
\begin{equation}
x^\mathrm{bc} = \mathcal{S}_{\theta_\mathrm{re}}(x^\mathrm{ra}, c_c),
\end{equation}
where $x^\mathrm{bc}$ denotes the bootstrapped clear-weather video, $c_c$ is the clear-weather prompt, and $\mathcal{S}_{\theta_\mathrm{re}}$ denotes the sampling process with the reverse-editing model.
Compared with pseudo-clear input $x^\mathrm{pc}$, $x^\mathrm{bc}$ better captures real-world effects such as sunlight glare and camera-motion blur.
We then pair it with the original real adverse-weather video $x^\mathrm{ra}$, forming a bootstrapped forward-editing pair $(x^\mathrm{bc}, x^\mathrm{ra})$.
We formulate the adapted forward training objective as:
\begin{equation}
\mathcal{L}_\mathrm{ad}
=
\mathbb{E}_{x^\mathrm{ra} \sim \mathcal{A}, \, \boldsymbol\epsilon, \, t}
[\|
f_\theta\left([\mathbf{z}^{\mathrm{ra}}_t,\mathbf{z}^{\mathrm{bc}}],t,c_a\right)
-
(\boldsymbol{\epsilon}-\mathbf{z}^{\mathrm{ra}})
\|_2^2],
\end{equation}
where $\mathbf{z}^\mathrm{bc} = \mathcal{E}(x^\mathrm{bc})$, $\mathbf{z}^\mathrm{ra} = \mathcal{E}(x^\mathrm{ra})$, and $\mathbf{z}^\mathrm{ra}_t = t\boldsymbol\epsilon + (1-t)\mathbf{z}^\mathrm{ra}$.
Finally, we jointly optimize the model with the reverse-editing and the real-world input adaptation objectives:
\begin{equation}
\mathcal{L}
=
\mathcal{L}_\mathrm{re}
+
\lambda_\mathrm{ad}\mathcal{L}_\mathrm{ad},
\end{equation}
where $\lambda_{\mathrm{ad}}$ balances the two directions. This joint training preserves the adverse-to-clear capability learned from real clear-weather targets, while refining clear-to-adverse translation with more realistic clear-style inputs. As a result, the model learns realistic weather effects from real-world videos, bridges the pseudo-to-real input gap, and naturally supports bidirectional clear~$\leftrightarrow$~adverse weather translation.

\subsection{Scene-Fidelity RL Optimization} \label{sec:reinforcement}

Although Progressive Realism Bootstrapping enables the model to learn realistic bidirectional weather translation, supervised fine-tuning alone may still introduce hallucinated objects or alter safety-critical scene structures. To explicitly enforce scene faithfulness, we introduce Scene-Fidelity RL Optimization, a reward-driven policy optimization strategy that optimizes the model with rule-based scene rewards.

\paragraph{Scene-Fidelity Reward.}
We design rewards to explicitly measure whether safety-critical driving elements are preserved after weather translation.
Given an input video $x$ and a translated video $\hat{x}$, we use SAM3~\cite{carion2025sam}, denoted by $\mathcal{M}$, to extract masks for key driving elements. Let $\mathcal{K}=\{\mathrm{lane}, \mathrm{ped}, \mathrm{veh}\}$ denote lane markings, pedestrians, and vehicles, with corresponding segmentation prompts $\{p_k\}_{k\in\mathcal{K}}$. The reward for each element is defined as
\begin{equation}
r_k = \mathrm{IoU}\big(\mathcal{M}(x,p_k), \mathcal{M}(\hat{x},p_k)\big),  \quad k \in \mathcal{K}.
\end{equation}
The final scene-fidelity reward is the weighted sum:
\begin{equation}
R_{\mathrm{scene}}
=
\sum_{k \in \mathcal{K}} w_k r_k,
\end{equation}
where $w_k$ controls the importance of each driving element.

\paragraph{Policy Optimization.}
We optimize our model with a GRPO-style objective~\cite{shao2024deepseekmath, liu2026flow}. In practice, we perform sampling and policy updates without text prompts, which is equivalent to setting classifier-free guidance~\cite{ho2022classifier} to zero during sampling. This design is motivated by two observations. First, the unconditional model tends to preserve the original weather appearance, making it suitable for evaluating scene-structure preservation. Second, computing IoU-based rewards between clear and adverse-weather videos is unreliable, since adverse weather can reduce visibility and change the detectability of lane markings, vehicles, and pedestrians. Under this setting, the segmentation-based rewards provide a clean signal for improving structural fidelity.
To prevent the model from losing its weather translation ability during policy optimization, we periodically interleave policy updates with supervised fine-tuning on the bootstrapped training pairs. This reward-driven optimization reduces hallucinations and improves the structural consistency of translated driving videos while maintaining the learned weather editing capability.

\subsection{Pseudo-Clear Generation} \label{sec:data-pipeline}

As discussed earlier, kickstarting our \textit{Progressive Realism Bootstrapping} requires an initial set of paired training data. Since collecting strictly paired clear and adverse weather videos of the same real-world driving scene is nearly infeasible, we design the auxiliary \textit{Pseudo-Clear Generation} pipeline $\mathcal{P}_{\mathrm{pc}}$ to construct reliable approximate counterparts. Additional implementation details of this pipeline are provided in the \underline{\textbf{Appendix}}.

Given a real adverse-weather video $x^\mathrm{ra}=\{x^\mathrm{ra}_t\}_{t=1}^{T}$, we edit its first frame $x^\mathrm{ra}_1$ into a pseudo-clear weather image $x^\mathrm{pc}_1$ using a state-of-the-art image editing model~\cite{flux-2-2025} while preserving the scene content. The edited image serves as a clear-style anchor frame. We then propagate this clear appearance to subsequent frames using a geometry-conditioned world model~\cite{ali2025world}, conditioned on per-frame geometric cues such as edge maps, depth maps, and semantic maps extracted from the video. This produces a temporally coherent pseudo-clear video $x^\mathrm{pc} = \{ x^\mathrm{pc}_t \}_{t=1}^{T}$. Additionally, to support text-conditioned flow matching, detailed weather captions $c_a$ for each adverse video $x^\mathrm{ra}$ are automatically generated using a vision-language model~\cite{bai2025qwen3}.

\paragraph{Weather-Agnostic Geometry Guidance.}
Geometric conditions extracted from adverse-weather videos often contain weather-specific structures, such as rain streaks and snowbanks, which can misguide pseudo-clear generation. To obtain weather-agnostic guidance, we first suppress transient precipitation effects using rain/snow removal models~\cite{lai2025snowmaster, jeong2025robust}, and then estimate geometric conditions from the cleaned video. We further use SAM3~\cite{carion2025sam} to mask out weather-dependent regions, such as puddles and snow-covered areas, from the geometry conditions while carefully preserving safety-critical driving elements (e.g., lane markings and vehicles).

\section{Experiments}

\subsection{Experimental Setup}

\paragraph{Datasets.}
Our evaluation set contains 100 clear-weather videos and 200 adverse-weather videos from the Waymo Open Dataset~\cite{Sun_2020_CVPR}, nuScenes~\cite{nuscenes2019}, and an internal driving dataset. The adverse-weather set covers fog, rain, and snow. Each clear video is translated into all three adverse-weather conditions, while all adverse-weather videos are used for adverse-to-clear evaluation.

\paragraph{Implementation Details.}
We initialize \method from Cosmos-Predict2.5-2B~\cite{ali2025world} and train it on 7,000 adverse-weather videos and 1,000 clear-weather videos. Each video contains 50 frames at 25 FPS and 720p resolution. Further preprocessing, optimization, and inference details are provided in the \underline{\textbf{Appendix}}.

\paragraph{Evaluation Metrics.}
We use CLIP Text--Image Similarity (\textbf{CS})~\cite{radford2021learning} for weather alignment, CLIP-Consistency (\textbf{CC})~\cite{instructnerf2023} for temporal consistency, DINO Structure Distance (\textbf{DS})~\cite{parmar2024one} for scene preservation, and PickScore (\textbf{PS})~\cite{Kirstain2023PickaPicAO} as a learned proxy for perceptual preference. We further introduce a VLM-based Realism score (\textbf{VR}), which uses real adverse-weather videos as in-context references to assess weather appearance and weather--scene interactions.
Higher values are better for CS, CC, PS, and VR, while lower values are better for DS.

\paragraph{Baselines.}
We compare with IntrinsicWeather~\cite{zhu2026intrinsicweather}, WeatherEdit~\cite{qian2026weatheredit}, AutoAWG~\cite{hu2026autoawg}, Cosmos-Transfer2.5~\cite{ali2025world}, SSGFormer~\cite{jeong2025robust}, and Wan2.1-VACE~\cite{vace}. These methods cover simulator-supervised translation, 3D weather editing, geometry-conditioned generation, adverse-weather removal, and general video editing.

\begin{figure}[t!]
    \centering
    \begin{subfigure}{\columnwidth}
        \centering
        \includegraphics[width=\columnwidth]{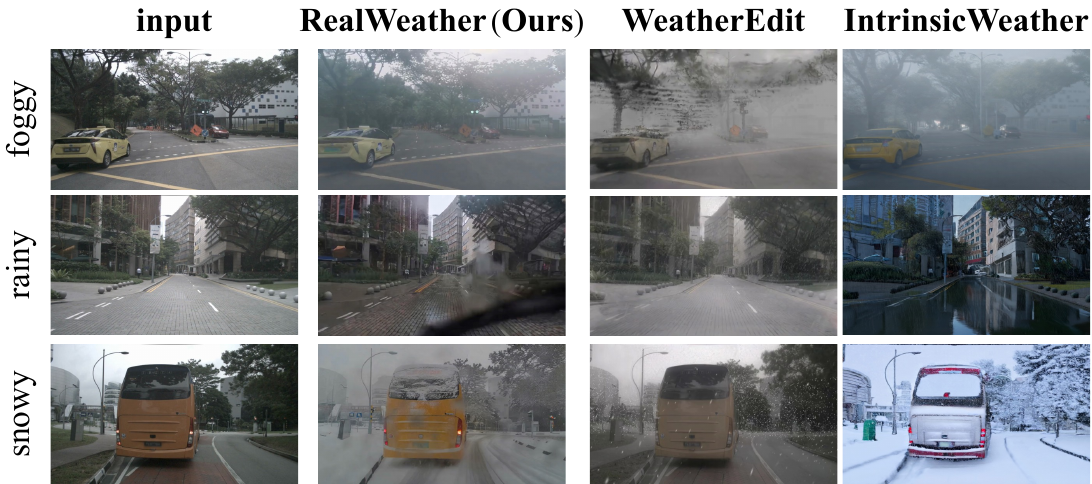}
        \caption{Clear-to-adverse weather synthesis.}
        \label{fig:cmp_top}
    \end{subfigure}
    \begin{subfigure}{\columnwidth}
        \centering
        \includegraphics[width=\columnwidth]{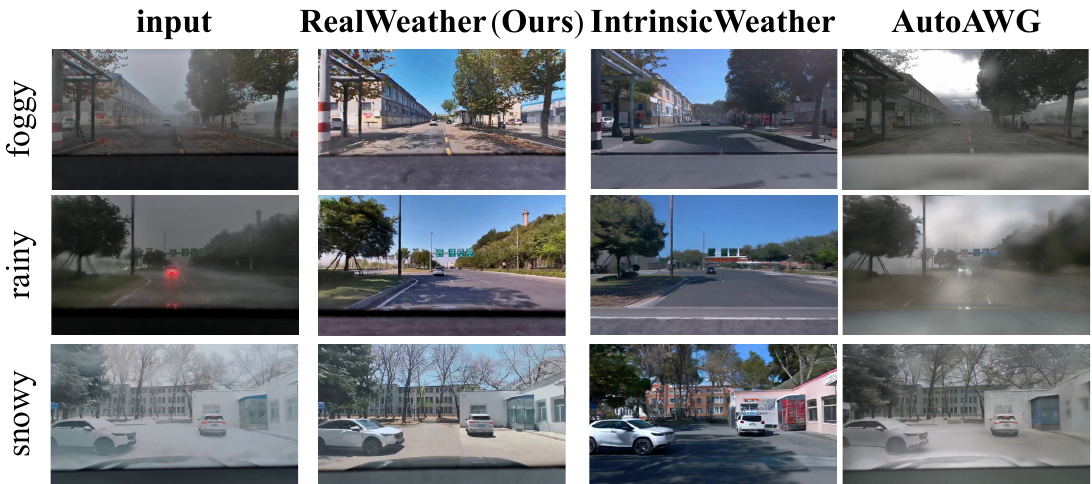}
        \caption{Adverse-to-clear weather removal.}
        \label{fig:cmp_bottom}
    \end{subfigure}
    \caption{
        Qualitative comparison with representative baselines.
        \method produces more realistic weather--scene interactions
        while preserving scene appearance in both translation directions.
    }
    \label{fig:cmp}
\end{figure}

\begin{table}[t!]
    \centering
    \begin{adjustbox}{width=\columnwidth}
        \setlength{\tabcolsep}{1.2mm}
        \begin{tabular}{lccccc}

        \multicolumn{6}{c}{\textbf{Clear-to-Adverse Translation}} \\
        \toprule
        \textbf{Method}
        & \textbf{CS} $\uparrow$
        & \textbf{CC} $\uparrow$
        & \textbf{DS} $\downarrow$
        & \textbf{PS} $\uparrow$
        & \textbf{VR} $\uparrow$ \\
        \cmidrule(lr){1-6}

        WeatherEdit
        & \best{0.23}
        & \second{0.90}
        & 0.07
        & 21.14
        & 3.63 \\

        IntrinsicWeather
        & \second{0.22}
        & -
        & 0.04
        & 21.34
        & 4.68 \\

        AutoAWG
        & \second{0.22}
        & 0.88
        & 0.03
        & \second{21.45}
        & \second{4.82} \\

        Cosmos-Transfer2.5-2B
        & 0.21
        & 0.88
        & \second{0.02}
        & 20.68
        & 4.16 \\

        Wan2.1-VACE-1.3B
        & 0.17
        & 0.83
        & \best{0.01}
        & 20.11
        & 2.24 \\

        \cmidrule(lr){1-6}
        \textbf{\method (Ours)}
        & \best{0.23}
        & \best{0.91}
        & 0.04
        & \best{21.52}
        & \best{4.99} \\

        \bottomrule
        \\
        \multicolumn{6}{c}{\textbf{Adverse-to-Clear Translation}} \\
        \toprule
        \textbf{Method}
        & \textbf{CS} $\uparrow$
        & \textbf{CC} $\uparrow$
        & \textbf{DS} $\downarrow$
        & \textbf{PS} $\uparrow$
        & \textbf{VR} $\uparrow$ \\
        \cmidrule(lr){1-6}

        IntrinsicWeather
        & 0.18
        & -
        & 0.05
        & 20.15
        & \second{4.29} \\

        AutoAWG
        & 0.19
        & \second{0.85}
        & \second{0.03}
        & 19.95
        & 1.51 \\

        Cosmos-Transfer2.5-2B
        & \best{0.21}
        & \best{0.88}
        & \second{0.03}
        & \second{20.38}
        & 2.77 \\

        Wan2.1-VACE-1.3B
        & \second{0.20}
        & 0.84
        & \best{0.01}
        & 19.63
        & 1.31 \\

        SSGFormer
        & 0.17
        & 0.74
        & \best{0.01}
        & 19.68
        & 1.02 \\

        \cmidrule(lr){1-6}
        \textbf{\method (Ours)}
        & \second{0.20}
        & \best{0.88}
        & \second{0.03}
        & \best{20.55}
        & \best{4.51} \\

        \bottomrule
        \end{tabular}
    \end{adjustbox}

    \caption{
        Quantitative comparison with representative baselines.
        Higher values are better for CS, CC, PS, and VR, while lower
        values are better for DS. \textbf{Bold} and \underline{underlined} values denote
        the best and second-best performing results.
    }
    \label{tab:cmp-combined}
\end{table}

\subsection{Main Results}

\paragraph{Quantitative Results.}
Table~\ref{tab:cmp-combined} summarizes both translation directions. For clear-to-adverse translation, \method achieves the best CC, PS, and VR scores and ties for the best CS. For adverse-to-clear translation, it obtains the best PS and VR, ties for the best CC, and ranks second in CS. These results demonstrate strong weather alignment, temporal consistency, perceptual quality, and realism in both directions.
Some baselines achieve lower DS by applying more conservative edits. Since successful translation requires both sufficient weather changes and scene preservation, DS should be considered jointly with the remaining metrics. Overall, \method achieves the strongest realism and perceptual quality while maintaining competitive structural fidelity.

\paragraph{Qualitative Results.}
Figure~\ref{fig:cmp} shows that simulation- and synthetic-supervision-based methods reproduce common precipitation patterns but often struggle with complex weather--scene interactions. In contrast, \method generates more natural windshield raindrops, wiper effects, water splashes, wet-road reflections, and snow accumulation.
Geometry-conditioned methods preserve coarse layouts but may lose vehicle colors, road textures, and distant details. By conditioning directly on the input video and learning from real driving data, \method better preserves these appearance cues. For adverse-to-clear translation, it removes weather artifacts while retaining object identities and scene layout, whereas several baselines leave residual precipitation or blurred structures. Additional comparisons are provided in the \underline{\textbf{Appendix}}, with additional video results given in \underline{\textbf{Project Page}}.

\subsection{Long-Tail and Out-of-Distribution Generalization}

\begin{figure}[t!]
    \centering
    \includegraphics[width=\columnwidth]{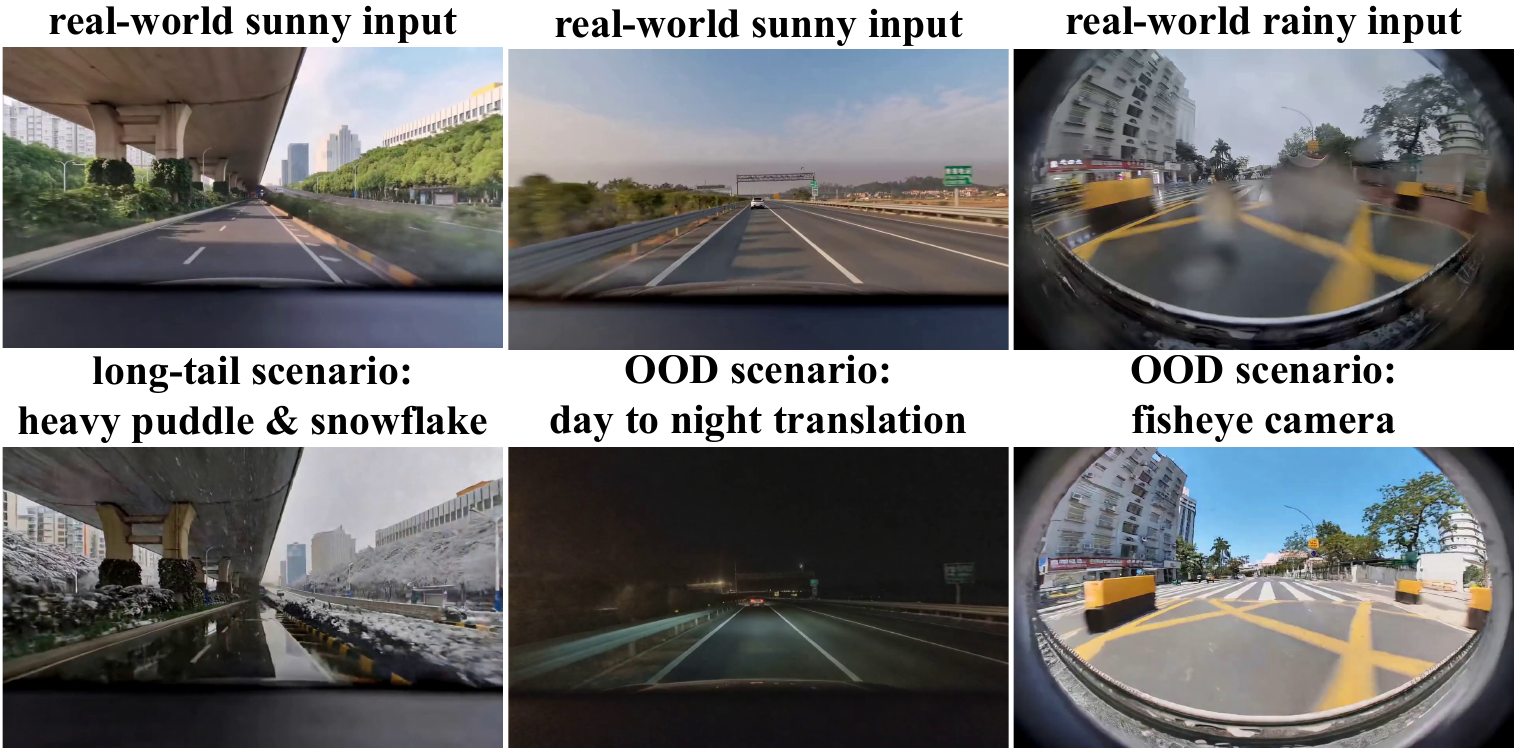}
    \caption{
        Qualitative results on long-tail and out-of-distribution inputs.
        \method handles compound weather, day-to-night appearance editing,
        and fisheye-camera videos without additional training.
    }
    \label{fig:ood}
\end{figure}

Beyond standard fog, rain, and snow, we evaluate \method on more challenging conditions. Figure~\ref{fig:ood} shows that the model can synthesize compound scenarios, such as dense snowfall with extensive road puddles, while preserving the main traffic scene. This requires consistent generation of atmospheric precipitation, ground-level effects, and scene-dependent reflections.
We also explore conditions outside the training scope. When prompted with nighttime descriptions, \method produces darker illumination and headlight-related effects. It further maintains reasonable temporal coherence on fisheye-camera videos despite the change in image geometry. These qualitative results suggest that the pretrained driving-world-model prior supports transfer beyond the training distribution.

\begin{figure}[t!]
    \centering

    \begin{subfigure}{0.85\columnwidth}
        \centering
        \begin{tikzpicture}
        \begin{axis}[
            ybar,
            width=\columnwidth,
            height=3.0cm,
            ymin=4.3, ymax=5.1,
            bar width=18pt,
            enlarge x limits=0.35,
            ylabel={VLM-Realism},
            symbolic x coords={w/o Bootstrapping, w/ Bootstrapping},
            xtick=data,
            xticklabel style={align=center,font=\small},
            yticklabel style={font=\small},
            ylabel style={font=\small},
            nodes near coords,
            every node near coord/.append style={font=\small},
            axis x line*=bottom,
            axis y line*=left,
            tick style={black},
        ]
        \addplot coordinates {
            (w/o Bootstrapping, 4.47)
            (w/ Bootstrapping, 4.99)
        };
        \end{axis}
        \end{tikzpicture}
        \caption{}
        \label{fig:abl-bootstrap-plot}
    \end{subfigure}

    \vspace{0.0em}

    \begin{subfigure}{0.85\columnwidth}
        \centering
        \begin{tikzpicture}
        \begin{axis}[
            ybar,
            width=\columnwidth,
            height=3.0cm,
            ymin=0.45, ymax=0.75,
            bar width=18pt,
            enlarge x limits=0.35,
            ylabel={SAM3-IoU},
            symbolic x coords={w/o Scene-Fidelity RL, w/ Scene-Fidelity RL},
            xtick=data,
            xticklabel style={align=center,font=\small},
            yticklabel style={font=\small},
            ylabel style={font=\small},
            nodes near coords,
            every node near coord/.append style={font=\small},
            axis x line*=bottom,
            axis y line*=left,
            tick style={black},
        ]
        \addplot coordinates {
            (w/o Scene-Fidelity RL, 0.54)
            (w/ Scene-Fidelity RL, 0.69)
        };
        \end{axis}
        \end{tikzpicture}
        \caption{}
        \label{fig:abl-rl-plot}
    \end{subfigure}

    \caption{
        Ablation results of the two main components in \method: (a) Effect of Progressive Realism Bootstrapping, and (b) Effect of Scene-Fidelity RL Optimization.
    }
    \label{fig:abl-main}
\end{figure}
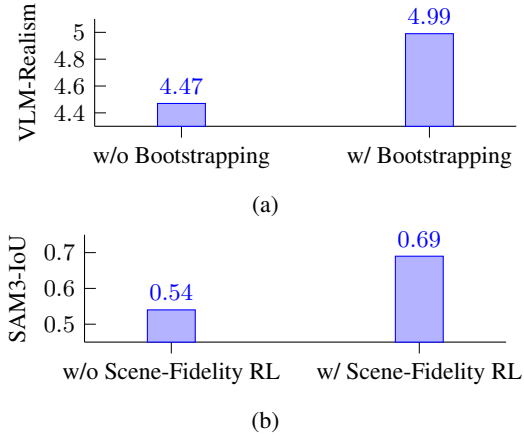

\subsection{Ablation Studies}

We isolate the effects of the two main training components. We first evaluate whether Progressive Realism Bootstrapping reduces the distribution gap between pseudo-clear and real clear-weather inputs, and then study whether Scene-Fidelity RL Optimization improves structural preservation.

\paragraph{Effect of Progressive Realism Bootstrapping.}
\begin{figure}[t!]
    \centering
    \includegraphics[width=\columnwidth]{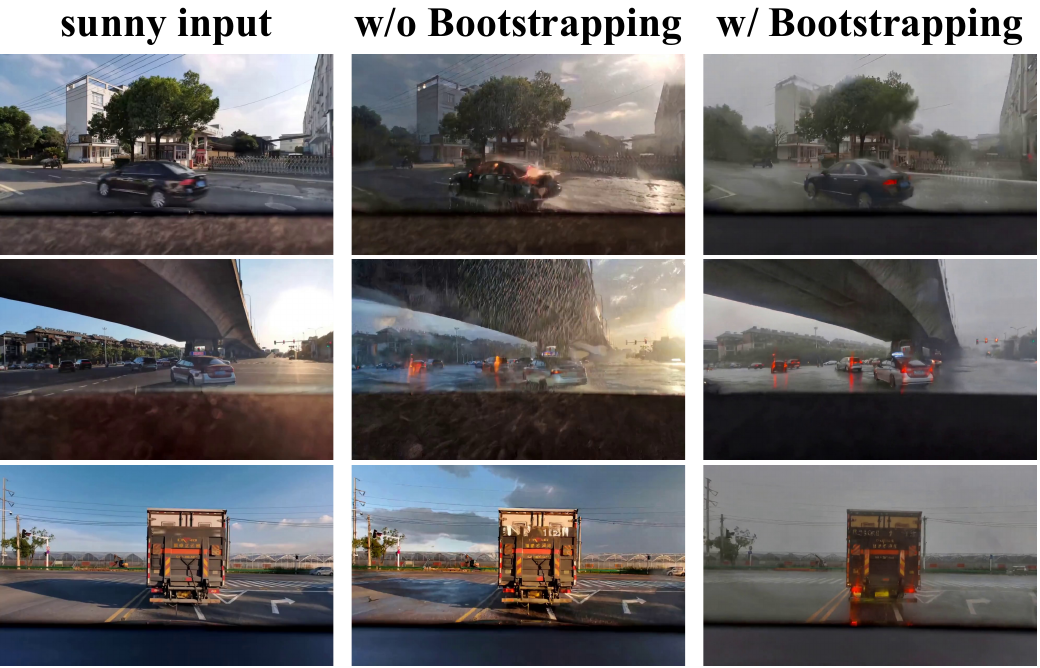}
    \caption{
        Effect of Progressive Realism Bootstrapping.
        Without bootstrapping, real-image glare can cause unnatural
        precipitation; bootstrapping improves adaptation to real inputs.
    }
    \label{fig:abl-bootstrapping}
\end{figure}

\begin{figure}[t!]
    \centering
    \includegraphics[width=\columnwidth]{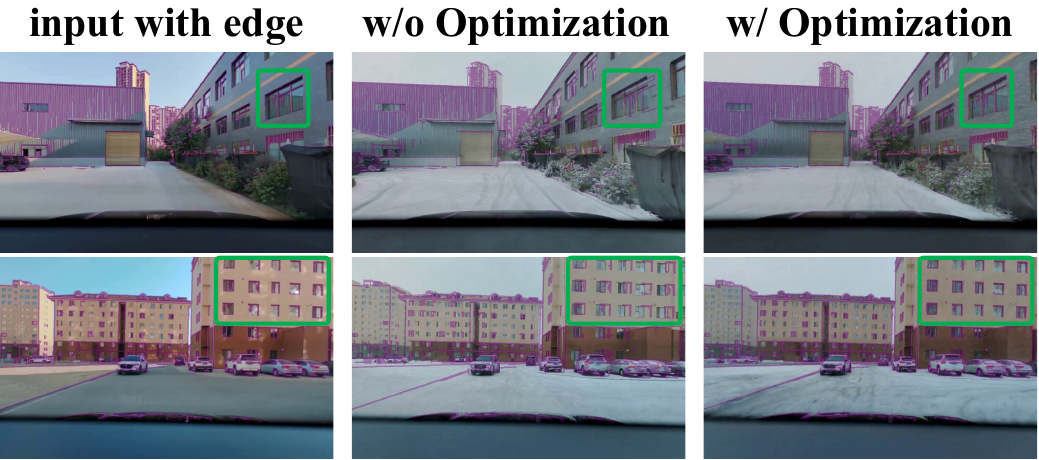}
    \caption{
        Effect of Scene-Fidelity RL Optimization.
        The optimized model reduces local deformation and better
        preserves object boundaries and road structures.
    }
    \label{fig:abl-rl}
\end{figure}

We compare the initial forward model, trained only on pseudo-clear-to-real-adverse pairs, with the complete bootstrapped model. Because the initial model observes only pseudo-clear conditioning inputs, it may not generalize to real clear videos containing sunlight glare, exposure variation, or motion blur.
As reported in Figure~\ref{fig:abl-main}(a), bootstrapping improves the VLM-Realism score from 4.47 to 4.99. Figure~\ref{fig:abl-rl} shows that, without bootstrapping, bright regions may be misinterpreted as weather-related structures, producing unnatural precipitation or weak weather changes. The bootstrapped model instead generates more consistent weather effects while better preserving the original appearance, supporting the importance of reducing the pseudo-to-real input gap.

\paragraph{Effect of Scene-Fidelity RL Optimization.}

We compare the bootstrapped model before and after Scene-Fidelity RL Optimization. Figure~\ref{fig:abl-main}(b) shows that the average SAM3-IoU increases from 0.54 to 0.69, indicating improved preservation of safety-critical scene elements.
The qualitative results in Figure~\ref{fig:abl-rl} are consistent with this gain. Before optimization, the model may introduce local deformation or modify object boundaries during weather synthesis. After optimization, the translated content better aligns with the input, with stronger correspondence to its Canny edges. This suggests that the reward reduces broader structural distortions in addition to improving the semantic elements explicitly included in the optimization.

\section{Conclusions}
We presented \method, a driving world model adapted for realistic and scene-faithful weather translation in driving videos.
To learn realistic weather effects without paired real-world clear/adverse videos, we introduced Progressive Realism Bootstrapping with a Pseudo-Clear Generation pipeline, enabling bidirectional clear~$\leftrightarrow$~adverse weather translation from real driving videos.
We further proposed Scene-Fidelity RL Optimization with segmentation-based rewards to reduce hallucinations and preserve key driving-scene structures.
Experiments demonstrate that \method outperforms simulation-based, geometry-conditioned, and general video editing baselines in both weather realism and scene fidelity. \method also supports long-tail weather scenarios and generalizes beyond the training distribution, including unseen camera settings such as fisheye views and novel transformations such as day-to-night translation.

\bibliography{aaai2027}

\input{appendix}

\end{document}

%% file: appendix.tex
\clearpage
\setcounter{page}{1}

\twocolumn[{
    \centering
    {\LARGE\bfseries \method: Realistic and Scene-Faithful Weather Translation with Driving World Models \\ (Appendix) \par}
    \vspace{1.6em}

    \includegraphics[width=\linewidth]{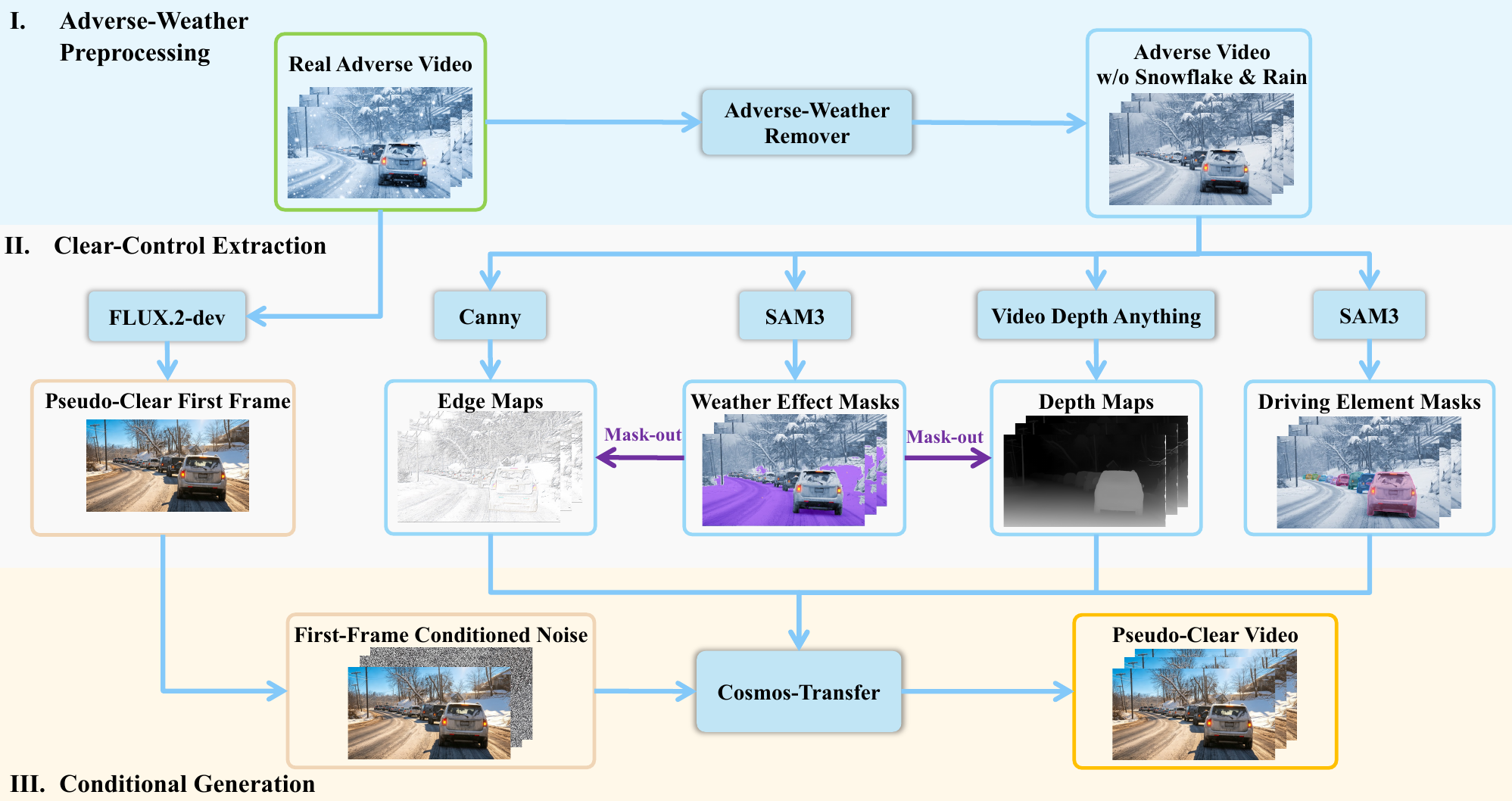}\par
    \captionof{figure}{
        Overview of the \textbf{Pseudo-Clear Generation} pipeline of \textbf{\method}.
        \textit{Adverse-Weather Preprocessing} removes transient precipitation artifacts using established adverse-weather removal models.
        \textit{Clear-Control Extraction} constructs a clear-weather anchor frame and weather-agnostic geometric controls. Finally, \textit{Conditional Generation} leverages a state-of-the-art driving world model to propagate the clear-weather appearance from the anchor frame throughout the video, producing a scene-consistent pseudo-clear counterpart.
    }
    \label{fig:datapipeline}
    \vspace{2em}
}]

\section{Appendix Overview}
This supplementary material complements the main paper with additional technical details, extended experimental results, and discussions of the limitations and future directions of \method. We organize the appendix as follows:
\begin{itemize}[leftmargin=*, itemsep=2pt]
    \item \textbf{Section A: Details of Pseudo-Clear Generation Pipeline.} We elaborate on the data preparation workflow necessary for kickstarting the bootstrapping process, including specific text prompts used for clear-weather image editing.
    \item \textbf{Section B: Experiment Details.} We provide extensive qualitative comparisons, implementation details regarding training hyperparameters and hardware, as well as a comprehensive breakdown of the VLM-Realism evaluation metric with exact prompts.
    \item \textbf{Section C: Limitations and Future Work.} We transparently discuss the current limitations of our framework, specifically concerning inference speed, precise weather-effect control, and generalization boundaries, alongside potential avenues for future research.
\end{itemize}

\section{A. Details of Pseudo-Clear Generation Pipeline}
\label{app:data-pipeline}

In this section, we elaborate on the structural components of our auxiliary \textit{Pseudo-Clear Generation} pipeline. This pipeline plays an essential role in constructing the initial paired training data necessary to kickstart our \textit{Progressive Realism Bootstrapping}. We first break down the step-by-step processing—from adverse-weather preprocessing to clear-control extraction and conditional generation—and subsequently detail the exact text prompts employed for image editing.

Figure~\ref{fig:datapipeline} illustrates the overall pipeline of \textbf{Pseudo-Clear Generation}. Given a real-world adverse-weather driving video, our goal is to generate a temporally coherent pseudo-clear counterpart that can serve as the input side of the initial supervision pairs for kickstarting the bootstrapping process. The pipeline consists of three main steps: \textbf{Adverse-Weather Preprocessing}, \textbf{Clear-Control Extraction}, and \textbf{Conditional Generation}.

\paragraph{Adverse-Weather Preprocessing.}
Directly extracting geometric conditions from adverse-weather videos can introduce weather-related artifacts into the control signals. In particular, transient weather effects such as falling rain streaks and snowflakes are high-frequency patterns that can strongly affect edge extraction. To reduce such interference, we first apply well-developed adverse-weather removal models~\cite{lai2025snowmaster, jeong2025robust} to suppress transient weather effects before computing geometry conditions. This preprocessing is only used to obtain cleaner control signals, rather than to generate the final pseudo-clear video.

\paragraph{Clear-Control Extraction.}
To guide pseudo-clear generation, we construct two complementary control signals: a \textit{Clear-Weather Anchor Frame}, which specifies the target clear-weather appearance, and \textit{Weather-Agnostic Geometry}, which preserves scene structure while suppressing weather-dependent cues. Together, they provide both appearance and structural guidance for propagating the clear style throughout the video.

\noindent\textit{Clear-Weather Anchor Frame.}
Given a real-world adverse-weather video, we use FLUX.2-dev~\cite{flux-2-2025} to translate its first frame into a clear-weather image while preserving the original scene structure. The resulting image serves as an anchor frame, providing an appearance reference for the subsequent video generation process.

\noindent\textit{Weather-Agnostic Geometry.}
To propagate the clear appearance across the remaining frames, we extract per-frame edge maps using the Canny operator and estimate depth maps with Video-Depth-Anything~\cite{video_depth_anything}. However, directly using these geometric signals is suboptimal because adverse-weather effects may introduce weather-dependent structures into both representations. For example, accumulated snow and road puddles can produce prominent edges and depth variations. Conditioning the driving world model on such structures may cause it to reproduce the corresponding adverse-weather effects, thereby weakening the pseudo-clear conversion.

To mitigate this adverse-to-clear geometry mismatch, we construct weather-agnostic geometric conditions. Specifically, we use SAM3~\cite{carion2025sam} with text prompts to segment weather-effect regions, such as snow-covered surfaces and puddles, and mask out the corresponding regions in the edge and depth maps. Directly suppressing weather-effect regions in edge and depth maps, however, may remove structural cues for safety-critical driving elements that overlap with adverse-weather effects, such as lane markings partially covered by snow. To compensate for this information loss, we further use SAM3 to segment key driving elements, including lane markings, vehicles, and pedestrians. The resulting semantic maps are introduced as an additional conditioning signal and jointly used with the weather-agnostic edge and depth maps to guide pseudo-clear generation.

\paragraph{Conditional Generation.}
Finally, we use a geometry-conditioned driving world model (Cosmos-Transfer2.5~\cite{ali2025world}) to propagate the clear style from the anchor frame to subsequent frames. The model takes the \textit{Clear-Weather Anchor Frame} and the \textit{Weather-Agnostic Geometry} as inputs, and generates the pseudo-clear video. To avoid unreliable appearance propagation caused by large viewpoint changes, we generate pseudo-clear videos in short clips of 2 seconds, corresponding to 50 frames at 25 FPS. Benefiting from the pretrained driving world model prior, our weather translation model can still generalize to longer video chunks during inference, including videos longer than 100 frames, and supports autoregressive long-video generation conditioned on previous frames, following the original driving world model.

\subsection{A.1. Implementation Details}
\paragraph{Prompts for FLUX.2-dev.}
We use FLUX.2-dev to edit the first frame of each input video and obtain the Clear-Weather Anchor Frame. To handle different weather conditions, we manually design editing prompts for \textit{cloudy}, \textit{foggy}, \textit{rainy}, and \textit{snowy} videos. These prompts remove weather-related appearance changes while preserving the underlying driving scene. We also introduce a separate real-clear-to-pseudo-clear prompt, which maps real-world clear-weather images into the pseudo-clear style used in our bootstrapping process. This prompt is used in the input-video bootstrapping stage to reduce the style gap between real-world clear inputs and pseudo-clear training inputs. The full prompt set is provided below.

\begin{promptbox}[Adverse-to-Clear Image Editing]
{\textbf{Prompt for Cloud Removal}}\\
Convert the scene to a dry, clear, cloudless, neutral-weather daytime environment. Absolutely remove all evidence of clouds. Remove all cloud coverage from the sky and reveal a natural, open sky. Preserve the original sky color and tone. Restore object original colors. Keep on-road elements and texture the same. Keep the car front hood and the dash top (cowl panel) exactly the same. Keep all scene geometry and content strictly identical, including road layout, vehicle shapes, building structures, and pedestrian positions.

\rule{\linewidth}{0.4pt}
{\textbf{Prompt for Fog Removal}}\\
Convert the scene to a dry, clear, neutral-weather daytime environment. Absolutely remove all evidence of fog. Remove any fog haze from the road surface. Remove any fog haze from sidewalks. Remove any fog haze from vehicles. Remove any fog haze from roofs and human-objects. Remove any fog haze from plants. Restore object original colors. Make roads and sidewalks fully visible without any low-visibility haze. Remove any fog haze in the air. Remove all cloud coverage from the sky and reveal a natural, open sky. Restore original colors of roads and sidewalks. Fix camera blur or occlusion caused by fog or mist on the lens. Keep on-road elements and texture the same. Keep the car front hood and the dash top (cowl panel) exactly the same. Keep all scene geometry and content strictly identical, including road layout, vehicle shapes, building structures, and pedestrian positions. Do not generate any new background buildings in areas previously obscured by fog. Leave background fogged regions as clean, neutral sky.

\rule{\linewidth}{0.4pt}
{\textbf{Prompt for Rain Removal}}\\
Convert the scene to a dry, clear, cloudless, neutral-weather daytime environment. Absolutely remove all evidence of rain. Remove any puddles from the road surface. Remove any puddles from sidewalks. Remove any rain streaks from vehicles. Remove any rain streaks from roofs and human-objects. Remove any rain streaks from plants. Restore object original colors. Make roads and sidewalks fully dry without any rain reflections or wet sheen. Restore original colors of roads and sidewalks. Remove any fog haze in the air. Remove all cloud coverage from the sky and reveal a natural, open sky. Fix camera blur or occlusion caused by raindrops or water on the lens. Keep on-road elements and texture the same. Keep the car front hood and the dash top (cowl panel) exactly the same. Keep all scene geometry and content strictly identical, including road layout, vehicle shapes, building structures, and pedestrian positions.

\rule{\linewidth}{0.4pt}
{\textbf{Prompt for Snow Removal}}\\
Convert the scene to a dry, clear, neutral-weather daytime environment. Absolutely remove all evidence of snow. Remove any snow residue from the road. Remove any snow residue from sidewalks. Remove any snow residue from vehicles. Remove any snow residue from roofs and human-objects. Remove any snow residue from plants. Remove any snow residue from trees. Remove any snow residue from bushes. Remove any puddles from the road surface. Remove any puddles from sidewalks. Make roads and sidewalks fully dry without any rain reflections or wet sheen. Remove any fog haze in the air. Remove all cloud coverage from the sky and reveal a natural, open sky. Restore object original colors. Fix camera blur or occlusion caused by snow or droplet on the lens. Keep on-road elements and texture the same. Keep the car front hood and the dash top (cowl panel) exactly the same. Keep all scene geometry and content strictly identical, including road layout, vehicle shapes, building structures, and pedestrian positions.

\rule{\linewidth}{0.4pt}
{\textbf{Prompt for Real-Clear-to-Pseudo-Style Editing}}\\
Convert the scene to a neutral-weather daytime environment with soft, even lighting. Absolutely remove all direct sun visibility from the sky. Remove all sunlight beams and glare. Remove all hard shadows cast on the road surface. Remove all hard shadows cast on sidewalks. Remove all hard shadows cast on vehicles, buildings, vegetation, and human-objects. Replace high-contrast lighting with diffuse, balanced illumination. Keep overall brightness gentle and not intense. Eliminate overexposed highlights and harsh contrast. Maintain a natural, clear atmosphere without dramatic lighting effects. Restore object original colors under neutral light. Keep on-road elements and texture the same. Keep the car front hood and the dash top (cowl panel) exactly the same. Keep all scene geometry and content strictly identical, including road layout, vehicle shapes, building structures, and pedestrian positions.
\end{promptbox}

\section{B. Experiment Details}

This section provides extended details regarding our experimental setup, qualitative results, and comprehensive evaluation metrics. We present additional visual comparisons demonstrating \method's superiority over existing baselines, detail the exact hyperparameters and hardware used during model training, and comprehensively outline the setup and exact prompts for our proposed novel VLM-Realism Score.

\subsection{B.1. Additional Experiment Results}
Additional qualitative results are provided on our \underline{\textbf{project website}}\footnote{Project page: \url{https://hust-umi.github.io/RealWeather/}}.
We include examples of single-weather translation, multi-weather translation, weather strength control, and adverse-weather removal. To evaluate generalization beyond the main test setting, we further show out-of-distribution results on driving-simulator videos, day-to-night editing, and fisheye-camera inputs. These examples demonstrate that \method can generalize to diverse weather combinations, unseen editing directions, and camera domain variations.

\subsection{B.2. Additional Qualitative Comparisons}
\begin{figure*}[ht!]
    \centering
    \includegraphics[width=0.70\textwidth]{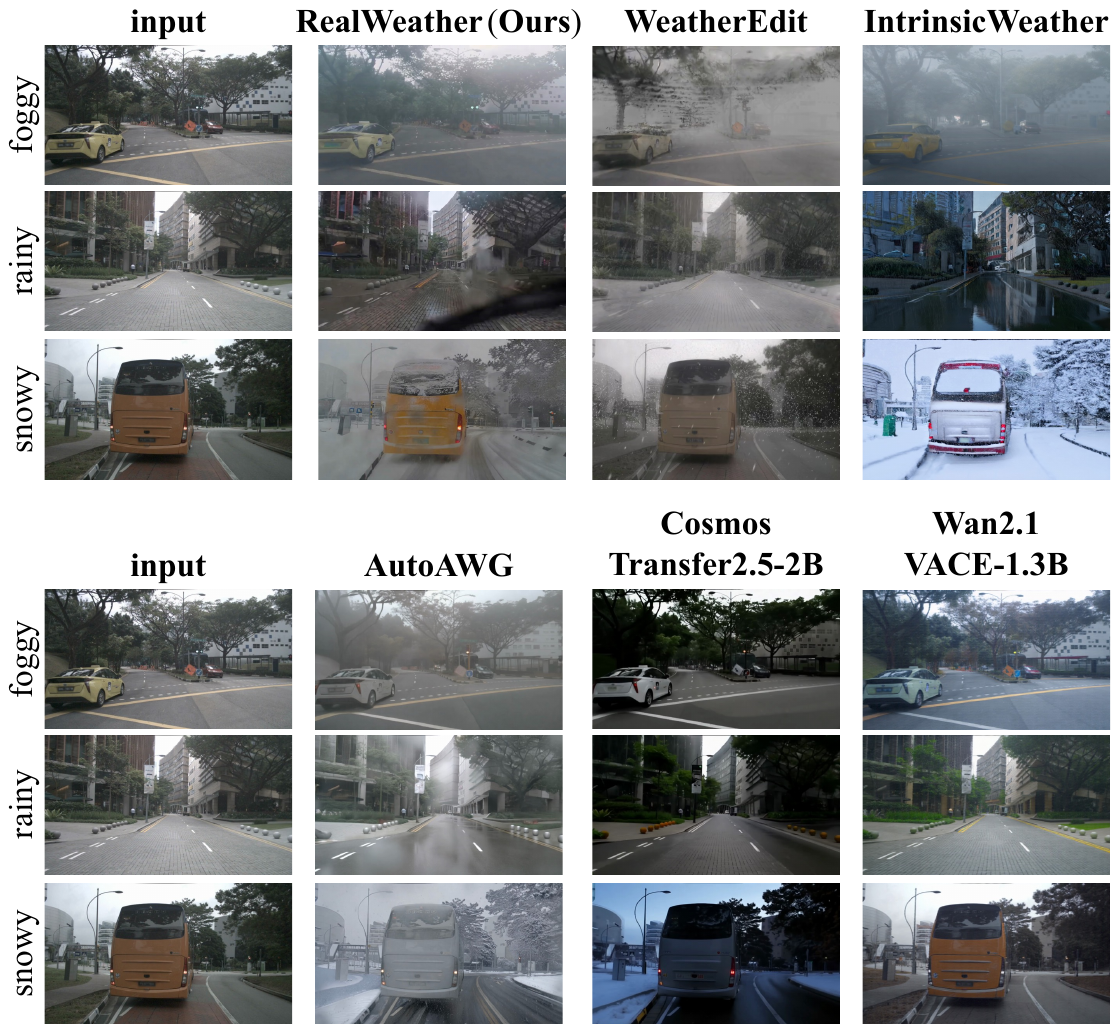}
    \caption{Comparison on clear-to-adverse translation with representative baselines, namely AutoAWG~\cite{hu2026autoawg}, Cosmos-Transfer2.5-2B~\cite{ali2025world}, Wan2.1-VACE-1.3B~\cite{vace}, WeatherEdit~\cite{qian2026weatheredit}, and InstrinsicWeather~\cite{zhu2026intrinsicweather}.}
    \label{fig:cmp-synthesis}
\end{figure*}

\begin{figure*}[ht!]
    \centering
    \includegraphics[width=0.70\textwidth]{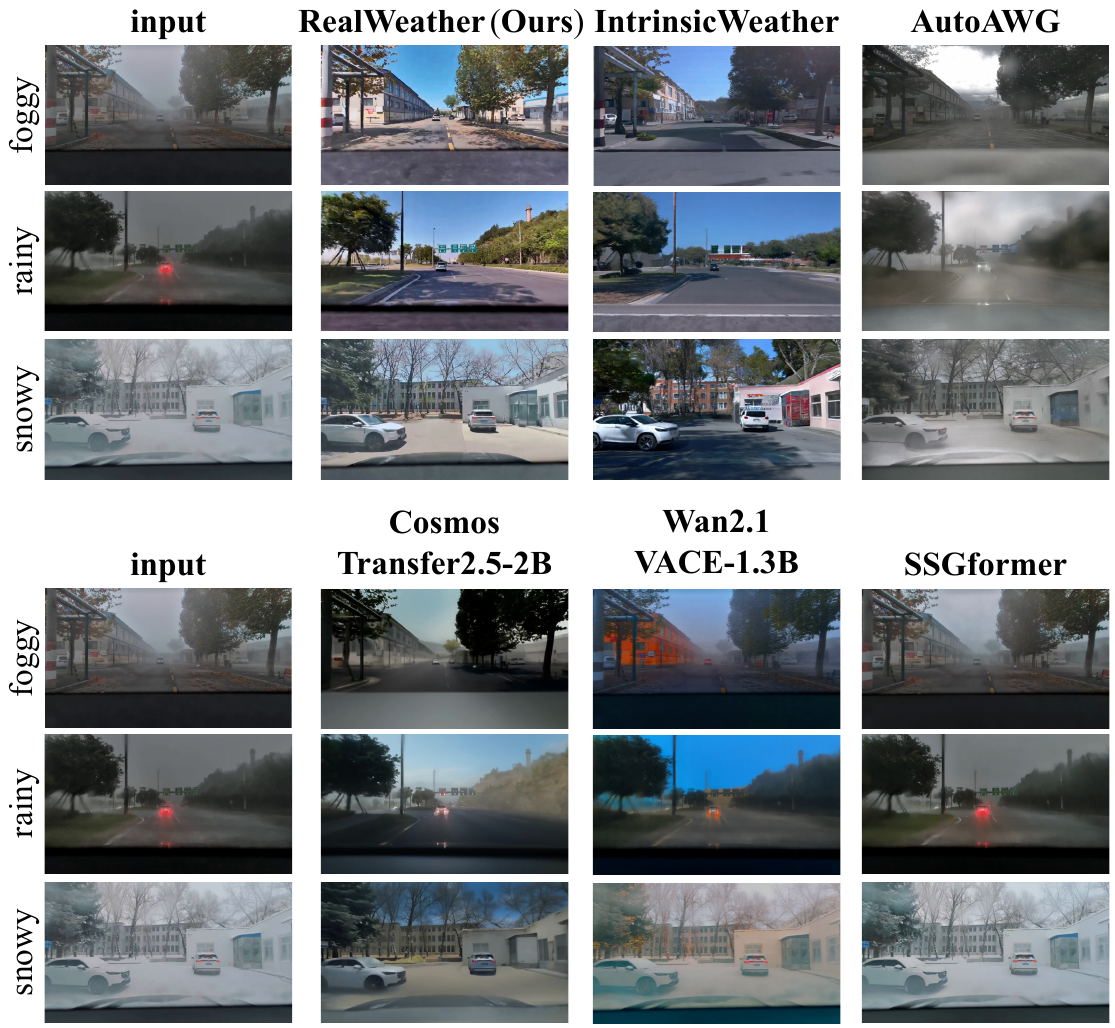}
    \caption{Comparison on adverse-to-clear translation with representative baselines, namely AutoAWG~\cite{hu2026autoawg}, Cosmos-Transfer2.5-2B~\cite{ali2025world}, Wan2.1-VACE-1.3B~\cite{vace}, SSGformer~\cite{jeong2025robust}, and InstrinsicWeather~\cite{zhu2026intrinsicweather}.}
    \label{fig:cmp-removal}
\end{figure*}
Qualitative comparisons against all baselines are presented in Figure~\ref{fig:cmp-synthesis} and Figure~\ref{fig:cmp-removal} for adverse-weather synthesis and removal, respectively. Consistent with the observations in the main paper, \method achieves superior weather realism and scene fidelity across both translation directions.

\paragraph{Comparison with synthetic-supervision / 3D-rendering methods.}
WeatherEdit and IntrinsicWeather rely on 3D-weather editing or training pairs from physics engines, which limits their ability to reproduce the realistic appearance and dynamics of real-world adverse weather. Their results often exhibit less natural precipitation patterns and weather-scene interactions. In contrast, \method learns weather effects directly from real-world video targets, producing more realistic weather appearance and dynamic phenomena, such as tire splashes and interactions between precipitation and the surrounding scene.

\paragraph{Comparison with geometry-conditioned models.}
AutoAWG and Cosmos-Transfer condition generation primarily on geometric representations such as edge and depth maps. Although these signals preserve coarse scene structure, they provide insufficient appearance information and may retain weather-dependent geometric cues introduced by puddles or accumulated snow. As a result, these models may produce geometry artifacts or lose fine-grained scene details. By combining appearance-aware conditioning with weather-agnostic structural guidance in our \textit{Pseudo-Clear Generation} pipeline, \method more effectively suppresses weather-induced artifacts while preserving road textures, vehicle appearance, and other important driving-scene structures. For adverse-to-clear translation, it also recovers sharper and more faithful clear-weather scenes than AutoAWG and Cosmos-Transfer.

\paragraph{Comparison with general-purpose video editing models.}
Wan2.1-VACE is a general-purpose video editing model that is not specifically adapted to driving-scene weather translation. It may introduce unintended changes to scene appearance or generate weather effects that insufficiently match the target condition. In contrast, \method builds upon a pretrained driving world model and learns weather translation from real-world driving videos, resulting in more realistic driving scenes, stronger adherence to the target weather condition, and better preservation of the original scene content.

\paragraph{Comparison with adverse-weather removal models.}
SSGFormer primarily focuses on removing transient precipitation artifacts, such as rain streaks and snowflakes, but often struggles with persistent scene-level weather effects, including puddles, snow accumulation, and weather-dependent illumination. \method handles both transient and persistent effects, producing clearer and more natural outputs while faithfully preserving the underlying driving scene.

\subsection{B.3. Implementation Details of \method}
We use Cosmos-Predict2.5-2B~\cite{ali2025world} as the pretrained driving world model. To adapt it for video-to-video weather translation, we extend the input channels to concatenate the noisy target latent with the input video latent, and initialize the newly added channels with zero convolution. During Progressive Realism Bootstrapping, we use 4,000 (pseudo-clear, real-adverse) pairs for Realistic Weather Effect Learning, 1,000 (realistic-adverse, real-clear) pairs for Reverse Weather Editing, and 3,000 (realistic-clear, real-adverse) pairs for Real-World Input Adaptation. The real-adverse pairs are balanced across rain, snow, and fog, with diverse weather intensities. The three training phases are optimized for 50k, 30k, and 50k steps, respectively, with a batch size of 8. All videos are in 720p resolution and contain 50 frames at 25 FPS. For Scene-Fidelity RL Optimization, we use the Flow-GRPO algorithm~\cite{liu2026flow}. At each RL step, we sample 8 independent groups, each containing 16 generated videos, and optimize the model for 400 steps. The reference policy is updated every 2 steps. All experiments are conducted on 8 NVIDIA B200 GPUs.

\subsection{B.4. Details of Evaluation Metrics}
We evaluate generated videos from five complementary perspectives.
For \textit{weather alignment}, we use CLIP Text-Image Similarity~\cite{radford2021learning} to measure the semantic alignment between generated frames and the target weather prompt.
For \textit{temporal consistency}, we report CLIP-Consistency~\cite{instructnerf2023}, which measures the directional similarity between pairs of original and edited videos in adjacent frames.
For \textit{scene faithfulness}, we adopt the DINO Structure score~\cite{parmar2024one} to measure structural preservation between the input and translated videos. 
For \textit{human preference}, we use PickScore~\cite{Kirstain2023PickaPicAO} to estimate the visual preference of generated results.
For \textit{real-world realism}, we further introduce a VLM-based evaluation metric, termed the \textbf{VLM-Realism Score}. Specifically, real driving videos are provided to the VLM~\cite{bai2025qwen3} as in-context references, and the model is asked to assess the realism of the generated weather videos. This evaluation complements prompt-based metrics by directly measuring their perceptual similarity to real-world driving videos.

\paragraph{Implementation of the VLM-Realism Score.}
We use Qwen3-VL-32B~\cite{bai2025qwen3} as the VLM evaluator to assess the realism of generated weather videos. Since standard prompt-alignment metrics mainly evaluate whether the output matches a textual weather description, they may not fully capture whether the generated video resembles real-world adverse-weather driving footage. To better evaluate real-world realism, we adopt an in-context learning protocol. Specifically, we provide the VLM with several real driving videos under different weather conditions as reference examples, and ask it to score the realism of generated videos.

For each evaluated video, the VLM is instructed to consider multiple aspects, including the naturalness of weather effects, temporal consistency, or any AI-generated flaws. The final VLM-Realism score is obtained by averaging the VLM scores over all evaluated videos. The prompt used for VLM evaluation is provided below.

\begin{promptbox}[Prompts for VLM-Realism Score]
{\large\textbf{Rain Video Evaluation}}\\
\textbf{Task Description:}\\
You evaluate if the video successfully changes a "Sunny Day" to a "Rainy Day" realistically. You provide a score from 1 to 5. You carefully check these key points for realism:
\begin{itemize}
    \item Overall Atmosphere \& Sky (The lighting is cloudy and the sky is overcast without bright sun.)
    \item Raindrops \& Road Water (Raindrops are visible and the road is wet with puddles and reflections.)
    \item Structures \& Temporal Continuity (Cars and buildings stay solid and the movement is smooth without shaking.)
    \item AI Flaws (The video is free of weird distortions, melting, or generative errors.)
\end{itemize}

\textbf{Evaluation Criteria}:\\
5 (Excellent): The video looks completely real. The overall atmosphere is dark and cloudy, and the sky is perfectly overcast. Raindrops are realistic, including those on the glass. The road has clear water, puddles, and accurate reflections. All structures stay solid. Temporal continuity is perfect with smooth movement and no shaking. The video has no AI flaws. \\
4 (Good): The video looks like a real rainy day. The overall atmosphere, sky, and road water look convincing. Structures are stable with good temporal continuity. The video contains very small AI flaws, but the physics and logic remain correct. \\
3 (Fair): Rain is added, but the video has big mistakes. The sky still shows a bright sun, or sharp sunny shadows are visible. The road looks dry with no water or reflections. Structures warp slightly, or the video shakes. AI flaws are noticeable. \\
2 (Poor): The quality is low. The video looks like a simple transparent rain filter placed on top of a sunny video. The sky, lighting, and atmosphere do not match a rainy day. Structures warp, bend, or melt. The video has a high amount of AI flaws. \\
1 (Bad): The video is a total failure. Heavy flickering occurs, and the video breaks down completely. Structures collapse, movement is broken without temporal continuity, and there is no rainy atmosphere at all. 

\rule{\linewidth}{0.4pt}
In-Context Examples:\\

[Example 1] \\
Input Video: \textless video\textgreater \\
Analysis: The overall atmosphere perfectly feels like a rainy day with a dark sky, realistic raindrops on the glass, and wet roads reflecting car lights. All structures remain perfectly stable with excellent temporal continuity and smooth movement. There are no AI flaws, as the side blur represents normal camera speed blur. \\
Score: 5 \\

[Example 2] \\
Input Video: \textless video\textgreater \\
Analysis: The overall realism of the video is relatively good. The sky is cloudy, the ground is wet and reflective, and the temporal continuity is acceptable. However, no rain is observed to have been added. \\
Score: 2 \\
\rule{\linewidth}{0.4pt}
Test Task: \\
You analyze the following uploaded video. You provide a brief analysis consisting of one or two sentences first, and you conclude with the final score in the exact format: `Final Score: [X]`. \\

Input Video: \textless video\textgreater \\
Analysis: \textless answer\textgreater\\ 
\\

{\large\textbf{Foggy Video Evaluation}}\\
\textbf{Task Description:}\\
You evaluate if the video successfully changes a "Sunny Day" to a "Foggy Day" realistically. You provide a score from 1 to 5. You carefully check these key points for realism:
\begin{itemize}
    \item Overall Atmosphere \& Sky (The lighting is diffused and the overall visibility is significantly reduced.)
    \item Fog Density \& Depth (The fog realistically obscures distant objects while close objects remain clearer, creating a sense of depth.)
    \item Structures \& Temporal Continuity (Cars and buildings stay solid and the movement is smooth without shaking.)
    \item AI Flaws (The video is free of weird distortions, melting, or generative errors.)
\end{itemize}

\textbf{Evaluation Criteria}:\\
5 (Excellent): The video looks completely real. The overall atmosphere accurately represents a foggy day. The fog exhibits realistic depth cueing, naturally hiding distant buildings while keeping the immediate road visible. All structures stay solid. Temporal continuity is perfect with smooth movement and no shaking. The video has no AI flaws.\\
4 (Good): The video looks like a real foggy day. The overall atmosphere and depth cueing look convincing. Structures are stable with good temporal continuity. The video contains very small AI flaws, but the physics and logic remain correct.\\
3 (Fair): Fog is added, but the video has big mistakes. The fog is rendered as a flat, uniform semi-transparent layer without any depth cueing. Structures warp slightly, or the video shakes. AI flaws are noticeable.\\
2 (Poor): The quality is low. The video looks like a simple white overlay filter placed on top of a sunny video. The lighting and atmosphere do not match a foggy day. Structures warp, bend, or melt. The video has a high amount of AI flaws.\\
1 (Bad): The video is a total failure. Heavy flickering occurs, the video breaks down completely, or the target weather is entirely missing (e.g., showing just a cloudy day with clear visibility and no fog).\\
\rule{\linewidth}{0.4pt}
In-Context Examples:\\

[Example 1] \\
Input Video: \textless video\textgreater \\
Analysis: The overall atmosphere perfectly feels like a foggy day with realistic depth cueing that naturally obscures distant buildings. All structures remain perfectly stable with excellent temporal continuity and smooth movement. There are no AI flaws. \\
Score: 5\\

[Example 2] \\
Input Video: \textless video\textgreater \\
Analysis: The video displays a cloudy sky and stable structures, but there is absolutely no fog present to reduce visibility. It completely fails to create a foggy environment, leaving distant objects fully visible. The video is merely a standard overcast scene.\\
Score: 1\\
\rule{\linewidth}{0.4pt}
Test Task: \\
You analyze the following uploaded video carefully. You provide a brief analysis consisting of one or two sentences first, and you conclude with the final score in the exact format: `Final Score: [X]`. \\

Input Video: \textless video\textgreater \\
Analysis: \textless answer\textgreater\\

{\large\textbf{Snowy Video Evaluation}}\\
\textbf{Task Description:}\\
You evaluate if the video successfully changes a "Sunny Day" to a "Snowy Day" realistically. You provide a score from 1 to 5. You carefully check these key points for realism:
\begin{itemize}
    \item Overall Atmosphere \& Sky (The lighting is diffused and the sky is overcast typical of a winter day.)
    \item Snowflakes \& Accumulation (Falling snowflakes are visible and snow realistically accumulates on the road, cars, and trees.)
    \item Structures \& Temporal Continuity (Cars and buildings stay solid and the movement is smooth without shaking.)
    \item AI Flaws (The video is free of weird distortions, melting, or generative errors.)
\end{itemize}

\textbf{Evaluation Criteria}:\\
5 (Excellent): The video looks completely real. The overall atmosphere is naturally overcast. Falling snow and geometric snow accumulation on surfaces look highly realistic. All structures stay solid. Temporal continuity is perfect with smooth movement and no shaking. The video has no AI flaws.\\
4 (Good): The video looks like a real snowy day. The overall atmosphere, sky, and snow accumulation look convincing. Structures are stable with good temporal continuity. The video contains very small AI flaws, but the physics and logic remain correct.\\
3 (Fair): Snow is added, but the video has big mistakes. The sky still shows a bright sun, or snow floats unnaturally without accumulating on the ground. Structures warp slightly, or the video shakes. AI flaws are noticeable.\\
2 (Poor): The quality is low. The video looks like a simple white filter or flat white paint mapped over a sunny video. The sky, lighting, and atmosphere do not match a snowy day. Structures warp, bend, or melt. The video has a high amount of AI flaws.\\
1 (Bad): The video is a total failure. Heavy flickering occurs, the video breaks down completely, or the target weather is entirely missing (e.g., showing just a cloudy day with absolutely no snow).\\
\rule{\linewidth}{0.4pt}
In-Context Examples:\\

[Example 1] \\
Input Video: \textless video\textgreater \\
Analysis: The overall atmosphere perfectly feels like a snowy day with a diffused sky, natural falling snowflakes, and realistic snow accumulation on the road. All structures remain perfectly stable with excellent temporal continuity and smooth movement. There are no AI flaws. \\
Score: 5 \\

[Example 2] \\
Input Video: \textless video\textgreater \\
Analysis: The video presents a cloudy atmosphere with acceptable temporal continuity, but it completely lacks any falling snowflakes or snow accumulation on the ground. It fails entirely to generate a snowy environment. The scene remains just a normal cloudy day.\\
Score: 1\\
\rule{\linewidth}{0.4pt}
Test Task: \\
You analyze the following uploaded video. You provide a brief analysis consisting of one or two sentences first, and you conclude with the final score in the exact format: `Final Score: [X]`. \\

Input Video: \textless video\textgreater \\
Analysis: \textless answer\textgreater\\

{\large\textbf{Sunny Video Evaluation}}\\
\textbf{Task Description:}\\
You evaluate if the video successfully changes a "Foggy/Snowy/Rainy Day" to a "Sunny Day" realistically. You provide a score from 1 to 5. You carefully check these key points for realism:
\begin{itemize}
    \item Overall Atmosphere \& Sky (The lighting is bright and the sky is predominantly clear or blue.)
    \item Sunlight \& Shadows (Strong, realistic directional sunlight is present, casting sharp and accurate shadows on the ground.)
    \item Structures \& Temporal Continuity (Cars and buildings stay solid and the movement is smooth without shaking.)
    \item AI Flaws (The video is free of weird distortions, melting, or generative errors.)
\end{itemize}

\textbf{Evaluation Criteria}:\\
5 (Excellent): The video looks completely real. The overall atmosphere is bright, and the sky is clear. Realistic directional sunlight casts sharp, physically accurate shadows. All structures stay solid. Temporal continuity is perfect with smooth movement and no shaking. The video has no AI flaws.\\
4 (Good): The video looks like a real sunny day. The overall atmosphere, sunlight, and shadows look convincing. Structures are stable with good temporal continuity. The video contains very small AI flaws, but the physics and logic remain correct.\\
3 (Fair): Sunlight is added, but the video has big mistakes. The sky looks artificial, or the cast shadows contradict the light source. Structures warp slightly, or the video shakes. AI flaws are noticeable.\\
2 (Poor): The quality is low. The video looks like a simple yellow/warm color filter placed on top of a dark video. The lighting lacks actual directional shadows. Structures warp, bend, or melt. The video has a high amount of AI flaws.\\
1 (Bad): The video is a total failure. Heavy flickering occurs, the video breaks down completely, or the target weather is entirely missing (e.g., still showing rain, wet roads, or heavy dark clouds).\\
\rule{\linewidth}{0.4pt}
In-Context Examples:\\

[Example 1] \\
Input Video: \textless video\textgreater \\
Analysis: The overall atmosphere perfectly feels like a sunny day with a bright blue sky and realistic sharp shadows cast by the vehicles. All structures remain perfectly stable with excellent temporal continuity and smooth movement. There are no AI flaws. \\
Score: 5 \\

[Example 2] \\
Input Video: \textless video\textgreater \\
Analysis: The video still shows dark heavy clouds, a wet road, and visible falling raindrops. It fails entirely to create a sunny atmosphere or produce any directional sunlight. The scene remains a rainy driving environment. \\
Score: 1 \\
\rule{\linewidth}{0.4pt}
Test Task: \\
You analyze the following uploaded video carefully. You provide a brief analysis consisting of one or two sentences first, and you conclude with the final score in the exact format: `Final Score: [X]`. \\

Input Video: \textless video\textgreater \\
Analysis: \textless answer\textgreater\\ 
\end{promptbox}

\section{C. Limitations and Future Work}

While \method achieves state-of-the-art realistic and scene-faithful weather translation, there remain practical areas for improvement. In this section, we transparently discuss the current limitations of our framework, specifically concerning inference speed, the precision of weather-effect control, and its generalization boundaries, alongside potential avenues for future research.

\paragraph{Inference Speed.}
The inference speed of generative driving world models is mainly affected by two factors: model scale and the number of sampling steps. Our model shares this limitation. As shown in Table~\ref{tab:inference-speed}, in the current implementation, we generate long videos at $1280\times720$ resolution in an autoregressive manner on a single node with 8 NVIDIA B200 GPUs. We use 35 sampling steps, and each inference chunk contains 101 frames, including one autoregressive conditioning frame. The overall throughput is about 2.5 frames per second, including VAE encoding/decoding, video encoding/decoding, and disk I/O. This speed is sufficient for offline data synthesis but still far from real-time deployment. Future work may improve efficiency by distilling the driving world model into a smaller model, reducing the number of sampling steps with consistency models or rectified-flow distillation, and optimizing the video decoding and storage pipeline.

\begin{table}[h]
\centering
\begin{tabular}{lc}
\toprule
Setting & Value \\
\midrule
Resolution & \(1280\times720\) \\
Sampling steps & 35 \\
Chunk length & 101 frames \\
Hardware & 8 NVIDIA B200 GPUs \\
Throughput & 2.5 FPS \\
\bottomrule
\end{tabular}
\caption{Inference speed of \method for long-video generation. The reported throughput includes VAE encoding/decoding, video encoding/decoding, and disk I/O.}
\label{tab:inference-speed}
\end{table}

\paragraph{Precise Weather-Effect Control.}
Currently, \method controls weather translation mainly through text prompts that describe weather type and intensity. While this enables flexible editing of rain, snow, fog, and mixed weather conditions, it does not yet provide fine-grained spatial control over weather effects. For example, users cannot precisely specify where snow should accumulate, whether snow should cover vehicles, trees, lanes, or sidewalks, or where puddles should appear on the road. A promising future direction is to develop more detailed weather captioning and control mechanisms.

\paragraph{Broader Real-World Condition Translation.}
Although \method is trained primarily for weather translation, it exhibits a certain degree of out-of-distribution generalization, such as translating daytime driving videos into nighttime scenes. This suggests the potential to extend driving world models beyond conventional weather conditions toward a broader range of real-world condition transformations. Examples include day-to-night translation and the generation of rare or extreme scenarios involving fires, floods, or other hazardous events. However, such transformations introduce substantial distribution shifts and require modeling complex changes in illumination, scene appearance, physical dynamics, and object--environment interactions. Extending \method to these extreme conditions while ensuring stable transformation, temporal consistency, and faithful preservation of the underlying driving scene remains an important direction for future research.